# Transfer Learning across Low-Resource, Related Languages for Neural Machine Translation

**Toan Q. Nguyen** and **David Chiang**
Department of Computer Science and Engineeering
University of Notre Dame
{tnguye28,dchiang}@nd.edu

## Abstract

We present a simple method to improve neural translation of a low-resource language pair using parallel data from a related, also low-resource, language pair. The method is based on the transfer method of Zoph et al., but whereas their method ignores any source vocabulary overlap, ours exploits it. First, we split words using Byte Pair Encoding (BPE) to increase vocabulary overlap. Then, we train a model on the first language pair and transfer its parameters, including its source word embeddings, to another model and continue training on the second language pair. Our experiments show that transfer learning helps word-based translation only slightly, but when used on top of a much stronger BPE baseline, it yields larger improvements of up to 4.3 BLEU.

## 1 Introduction

Neural machine translation (NMT) (Sutskever et al., 2014; Bahdanau et al., 2015) is rapidly proving itself to be a strong competitor to other statistical machine translation methods. However, it still lags behind other statistical methods on very low-resource language pairs (Zoph et al., 2016; Koehn and Knowles, 2017).

A common strategy to improve learning of low-resource languages is to use resources from related languages (Nakov and Ng, 2009). However, adapting these resources is not trivial. NMT offers some simple ways of doing this. For example, Zoph et al. (2016) train a *parent* model on a (possibly unrelated) high-resource language pair, then use this model to initialize a *child* model which is further trained on a low-resource language pair. In particular, they showed that a French-English model could be used to improve translation on a wide range of low-resource language pairs such as Hausa-, Turkish-, and Uzbek-English.

In this paper, we explore the opposite scenario, where the parent language pair is also low-resource, but related to the child language pair. We show that, at least in the case of three Turkic languages (Turkish, Uzbek, and Uyghur), the original method of Zoph et al. (2016) does not always work, but it is still possible to use the parent model to considerably improve the child model.

The basic idea is to exploit the relationship between the parent and child language lexicons. Zoph et al.'s original method makes no assumption about the relatedness of the parent and child languages, so it effectively makes a random assignment of the parent-language word embeddings to child-language words. But if we assume that the parent and child lexicons are related, it should be beneficial to transfer source word embeddings from parent-language words to their child-language equivalents.

Thus, the problem amounts to finding a representation of the data that ensures a sufficient overlap between the vocabularies of the languages. To do this, we map the source languages to a common alphabet and use Byte Pair Encoding (BPE) (Sennrich et al., 2016) on the union of the vocabularies to increase the number of common subwords.

In our experiments, we show that transfer learning helps word-based translation, but not always significantly. But when used on top of a much stronger BPE baseline, it yields larger and statistically significant improvements. Using Uzbek as a parent language and Turkish and Uyghur as child languages, we obtain improvements over BPE of 0.8 and 4.3 BLEU, respectively.

| English | Turkish | Uzbek |
|---|---|---|
| clinic | poliklinikte, poliklinie, polikliniine | poliklinikasi, poliklinikaga, poliklinikalar |
| parliament | parlamentosuna, parlamentosundan, parlamentosu | parlamentning, parlamentini, parlamentiga |
| meningococcus | meningokokuna, meningokosemi, meningokoklar | meningokokk, meningokokkli, meningokokkning |

Table 1: Some examples of similar words in Turkish and Uzbek that share the same root.

## 2 Background

### 2.1 Attentional Model

We use the 2-layer, 512-hidden-unit *global* attentional model with *general* scoring function and *input feeding* by Luong et al. (2015). For the purposes of this paper, the most important detail of the model is that (as in many other models) the word types of both the source and target languages are mapped to vector representations called *word embeddings*, which are learned automatically with the rest of the model.

### 2.2 Language transfer

We follow the transfer learning approach proposed by Zoph et al. (2016). In their work, a parent model is first trained on a high-resource language pair. Then the child model's parameter values are copied from the parent's and are fine-tuned on its low-resource data.

The source word embeddings are copied with the rest of the model, with the *i*th parent-language word embedding being assigned to the *i*th child-language word. Because the parent and child source languages have different vocabularies, this amounts to randomly assigning parent source word embeddings to child source words. In other words, even if a word exists in both the parent and child vocabularies, it's extremely unlikely that it will be assigned the same embedding in both models.

By contrast, because the target language is the same in both the parent and child models, the target word embeddings are frozen during fine-tuning.

### 2.3 Related languages

The experiments described below are on translation from three Turkic languages to English. The Turkic language family is a group of related languages with a very wide geographic distribution, from Turkey to northeastern Siberia. Turkic languages are morphologically rich, and have similarities in phonology, morphology, and syntax. For instance, in our analysis of the training data, we find many Turkish and Uzbek words sharing the same root and meaning. Some examples are shown in Table 1.

| model | word-based toks $\times 10^6$ | word-based sents $\times 10^3$ | BPE 5k toks $\times 10^6$ | BPE 5k sents $\times 10^3$ | BPE 60k toks $\times 10^6$ | BPE 60k sents $\times 10^3$ |
|---|---|---|---|---|---|---|
| Uzb par | 1.5 | 102 | 2.4 | 92 | 1.9 | 103 |
| Tur chd | 0.9 | 56 | 1.5 | 50 | 1.2 | 57 |
| Uzb par | 1.5 | 102 | 2.4 | 90 | 2.0 | 103 |
| Uyg chd | 1.7 | 82 | 2.1 | 77 | 2.0 | 88 |

Table 2: Number of tokens and sentences in our training data.

### 2.4 Byte Pair Encoding

BPE (Sennrich et al., 2016) is an efficient word segmentation algorithm. It initially treats the words as sequences of character tokens, then iteratively finds and merges the most common token pair into one. The algorithm stops after a controllable number of operations, or when no token pair appears more than once. At test time, the learned merge operations are applied to merge the character sequences in test data into larger symbols.

## 3 Method

The basic idea of our method is to extend the transfer method of Zoph et al. (2016) to share the parent and child's source vocabularies, so that when source word embeddings are transferred, a word that appears in both vocabularies keeps its embedding. In order for this to work, it must be the case that the parent and child languages have considerable vocabulary overlap, and that when a word occurs in both languages, it often has a similar meaning in both languages. Thus, we need to process the data to make these two assumptions hold as much as possible.

### 3.1 Transliteration

If the parent and child language have different orthographies, it should help to map them into a common orthography. Even if the two use the same script, some transformation could be applied; for example, we might change French *-eur* endings to Spanish *-or*. Here, we take a minimalist approach. Turkish and Uzbek are both written using Latin script, and we did not apply any transformations to them. Our Uyghur data is written in Arabic script, so we transliterated it to Latin script using an off-the-shelf transliterator.[1] The transliteration is a string homomorphism, replacing Arabic letters with English letters or consonant clusters independent of context.

### 3.2 Segmentation

To increase the overlap between the parent and child vocabularies, we use BPE to break words into subwords. For the BPE merge rules to not only find the common subwords between two source languages but also ensure consistency between source and target segmentation among each language pair, we learn the rules from the union of source and target data of both the parent and child models. The rules are then used to segment the corpora. It is important to note that this results in a single vocabulary, used for both the source and target languages in both models.

## 4 Experiments

We used Turkish-, Uzbek-, and Uyghur-English parallel texts from the LORELEI program. We tokenized all data using the Moses toolkit (Koehn et al., 2007); for Turkish-English experiments, we also truecased the data. For Uyghur-English, the word-based models were trained in the original Uyghur data written in Arabic script; for BPE-based systems, we transliterated it to Latin script as described above.

For the word-based systems, we fixed the vocabulary size and replaced out-of-vocabulary words with _UNK. We tried different sizes for each language pair; however, each word-based system's target vocabulary size is limited by that of the child, so we could only use up to 45,000 word types for Turkish-English and 20,000 for Uyghur-English.

The BPE-based systems could make use of bigger vocabulary size thanks to the combination of both parent and child source and target vocabularies. We varied the number of BPE merge operations from 5,000 to 60,000. Instead of using a fixed vocabulary cutoff, we used the full vocabulary; to ensure the model still learns how to deal with unknown words, we trained on two copies of the training data: one unchanged, and one in which all rare words (whose frequency is less than 5) were replaced with _UNK. Accordingly, the number of epochs was halved.

Following common practice, we fixed an upper limit on training sentence length (discarding longer sentences). Because the BPE-based systems have shorter tokens and therefore longer sentences, we set this upper limit differently for the word-based and BPE-based systems to approximately equalize the total size of the training data. This led to a limit of 50 tokens for word-based models and 60 tokens for BPE-based models. See Table 2 for statistics of the resulting datasets.

We trained using Adadelta (Zeiler, 2012), with a minibatch size of 32 and dropout with a dropout rate of 0.2. We rescaled the gradient when its norm exceeded 5. For the Uzbek-English to Turkish-English experiment, the parent and child models were trained for 100 and 50 epochs, respectively. For the Uzbek-English to Uyghur-English experiment, the parent and child models were trained for 50 and 200, respectively. As mentioned above, the BPE models were trained for half as many epochs because their data is duplicated.

We used beam search for translation on the dev and test sets. Since NMT tends to favor short translations (Cho et al., 2014), we use the length normalization approach of Wu et al. (2016) which uses a different score $s(e \mid f)$ instead of log-probability:

$$s(e \mid f) = \frac{\log p(e \mid f)}{lp(e)}$$

$$lp(e) = \frac{(5 + |e|)^{\alpha}}{(5 + 1)^{\alpha}}.$$

We set $\alpha = 0.8$ for all of our experiments.

[1] https://cis.temple.edu/~anwar/code/latin2uyghur.html

| | | baseline | | transfer | | transfer+freeze | |
|---|---|---|---|---|---|---|---|
| | | BLEU | size | BLEU | size | BLEU | size |
| Tur-Eng | word-based | 8.1 | 30k | 8.5* | 30k | 8.6* | 30k |
| | BPE | 12.4 | 10k | 13.2† | 20k | — | — |
| Uyg-Eng | word-based | 8.5 | 15k | 10.6† | 15k | 8.8* | 15k |
| | BPE | 11.1 | 10k | 15.4‡ | 8k | — | — |

Table 3: Whereas transfer learning at word-level does not always help, our method consistently yields a significant improvement over the stronger BPE systems. Scores are case-sensitive **test** BLEU. Key: `size` = vocabulary size (word-based) or number of BPE operations (BPE). The symbols † and ‡ indicate statistically significant improvements with $p < 0.05$ and $p < 0.01$, respectively, while $*$ indicates a statistically insignificant improvement ($p > 0.05$).

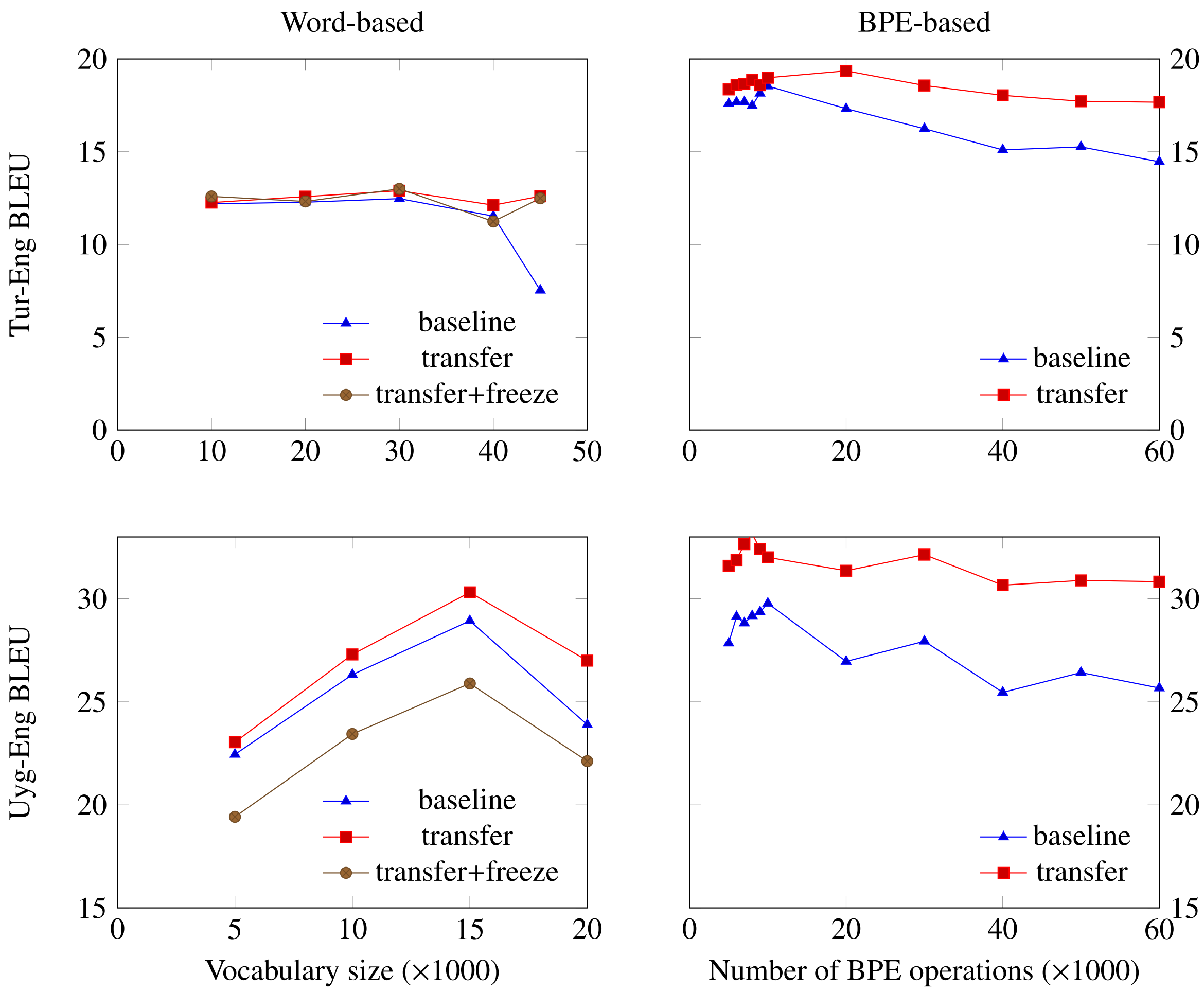

Figure 1: Tokenized **dev** BLEU scores for various settings as a function of the number of word/subword types. Key: baseline = train child model only; transfer = train parent, then child model; +freeze = freeze target word embeddings in child model.

| task | settings | train | dev |
|---|---|---|---|
| Tur-Eng | word-based 30k | 3.9% | 3.6% |
| | BPE 20k | 58.8% | 25.0% |
| Uyg-Eng | word-based 15k | 0.5% | 1.7% |
| | BPE 8k | 57.2% | 48.5% |

Table 4: Amount of child's source types that appear in parent.

We stopped training when the tokenized BLEU score was maximized on the development set. We also optimized the vocabulary size and the number of BPE operations for the word-based and BPE-based systems, respectively, to maximize the tokenized BLEU on the development set.

After translation at test time, we rejoined BPE segments, recased, and detokenized. Finally, we evaluated using case-sensitive BLEU.

As a baseline, we trained a child model using BPE but without transfer (that is, with weights randomly initialized). We also compared against a word-based baseline (without transfer) and two word-based systems using transfer without vocabulary-sharing, corresponding with the method of Zoph et al. (2016) (§2.2): one where the target word embeddings are fine-tuned, and one where they are frozen.

## 5 Results and Analysis

Our results are shown in Table 3. In this low-resource setting, we find that transferring word-based models does not consistently help. On Turkish-English, both transfer methods give only a statistically insignificant improvement ($p > 0.05$); on Uyghur-English, transfer without freezing target embeddings helps somewhat, but transfer with freezing helps only insignificantly.

In both language pairs, the models that use BPE perform much better than their word-based counterparts. When we apply transfer learning to this much stronger baseline, we find that the relative improvements actually increase; that is, the combined effect of BPE and transfer learning is more than additive. On Turkish-English, the improvement is +0.8 BLEU over the BPE baseline; on Uyghur-English, a healthy +4.3 over the BPE baseline.

A similar pattern emerges when we examine the best BLEU scores on the development set (Figure 1). Whereas word-based transfer methods help very little for Turkish-English, and help or hurt slightly for Uyghur-English, our BPE-based transfer approach consistently improves over both the baseline and transfer word-based models. We surmise that the improvement is primarily due to the vocabulary overlap created by BPE (see Table 4).

## 6 Conclusion

In this paper, we have shown that the transfer learning method of Zoph et al. (2016), while appealing, might not always work in a low-resource context. However, by combining it with BPE, we can improve NMT performance on a low-resource language pair by exploiting its lexical similarity with another related, low-resource language. Our results show consistent improvement in two Turkic languages. Our approach, which relies on segmenting words into subwords, seems well suited to agglutinative languages; further investigation would be needed to confirm whether our method works on other types of languages.

## 7 Acknowledgements

This research was supported in part by University of Southern California subcontract 67108176 under DARPA contract HR0011-15-C-0115. Nguyen was supported by a fellowship from the Vietnam Education Foundation. We would like to express our great appreciation to Dr. Sharon Hu for letting us use her group's GPU cluster (supported by NSF award 1629914), and to NVIDIA corporation for the donation of a Titan X GPU.